\documentclass[runningheads]{llncs}
\usepackage[T1]{fontenc}
\usepackage[utf8]{inputenc}
\usepackage[english]{babel}
\usepackage{csquotes}
\usepackage{mathtools,amsmath,dsfont}
\usepackage{graphicx}
\graphicspath{{./images/}}
\usepackage[inline]{enumitem}
\usepackage[%
table-number-alignment = center,
table-figures-integer = 1,
table-figures-decimal = 3]{siunitx}
\usepackage{xcolor,hyperref}
\hypersetup{
  pdfauthor={Gesina Schwalbe},
  pdftitle={Verification of Size Invariance in DNN Activations using Concept Embeddings},
}

\usepackage{booktabs,tabulary}

\addto\extrasenglish{}
\addto\extrasenglish{}

\newcommand*{\forexample}{e.g.\ }
\newcommand*{\idest}{i.e.\ }
\newcommand*{\concept}[1]{\textsf{#1}}

\begin{document}
 
\title{
  Verification of Size Invariance in DNN Activations using Concept Embeddings%
  \thanks{
    The research leading to these results was partly funded by the
    German Federal Ministry for Economic Affairs and Energy within the
    project \enquote{\href{https://www.ki-absicherung-projekt.de}{KI-Absicherung}}.
  }}
\titlerunning{Verification using Concept Embeddings}
\author{Gesina Schwalbe\inst{1,2}\orcidID{0000-0003-2690-2478} 
}
\authorrunning{G. Schwalbe}
\institute{
  Continental AG, Regensburg, Germany\\
  \email{gesina.schwalbe@continental-corporation.com}
  \and
  Cognitive Systems, University of Bamberg, Bamberg, Germany
}
\maketitle

\begin{abstract} 
  The benefits of deep neural networks (DNNs) have become of interest
  for safety critical applications like medical ones or automated
  driving.
  Here, however, quantitative insights into the DNN inner
  representations are mandatory~\cite{iso/tc22/sc32_iso_2018a}.
  One approach to this is \emph{concept analysis}, which aims to
  establish a mapping between the internal representation of a DNN and
  intuitive semantic concepts.
  Such can be sub-objects like human body parts that are
  valuable for validation of pedestrian detection.
  To our knowledge, concept analysis has not yet been applied to
  large object detectors, specifically not for sub-parts.
  Therefore, this work first suggests a substantially improved version
  of the Net2Vec approach~\cite{fong_net2vec_2018} for post-hoc
  segmentation of sub-objects.
  Its practical applicability is then demonstrated on a new concept
  dataset by two exemplary assessments of three standard networks,
  including the larger Mask~R-CNN model~\cite{he_mask_2017}:
  (1)~the consistency of body part similarity,
  and
  (2)~the invariance of internal representations of body parts with
  respect to the size in pixels of the depicted person.
  The findings show that the representation of body parts is mostly
  size invariant, which may suggest an early intelligent fusion of
  information in different size categories.
  
  \keywords{Concept embedding analysis \and MS~COCO \and Explainable AI.}
\end{abstract}


\section{Introduction} 
The high performance and flexibility of deep neural networks (DNNs)
makes them interesting for many complex computer vision applications.
This includes ethically involved or safety critical fields
where silent biases can become a matter of unfairness, or even
life-threatening decisions.
Hence, responsible artificial intelligence is required, allowing
for thorough assessments of trained
DNNs~\cite{arrieta_explainable_2020}.
For example in the area of automated driving,
quantitative insights into the inner representation
of trained DNNs are mandatory \cite{iso/tc22/sc32_iso_2018a}.

One step towards this direction is the research area of
\emph{concept (embedding) analysis}
\cite{fong_net2vec_2018,kim_interpretability_2018},
that started with the work in \cite{bau_network_2017} in 2017.
Supervised concept analysis aims to answer the question how (well)
information on a visual semantic concept is embedded in the
intermediate output of one layer of a trained DNN.
A concept can be, e.g., a texture, material, scene, object, or
object part \cite{bau_network_2017},
and should be given as samples with binary annotations.
The question is answered by training a simple model to
predict the concept from the layer output,
with simplicity ensuring interpretability
\cite{bau_network_2017,kim_interpretability_2018}.
Once these \emph{concept models} are obtained, several
means for verification open up:
The embedding quality can be measured as the prediction quality of the
concept model;
if a similarity measure for concept models is available,
semantic relations between concepts can be validated
(\forexample legs are similar to arms) \cite{fong_net2vec_2018};
and lastly, if the information embedding can be represented as a
vector in the latent space, one can use sensitivity
analysis to investigate dependencies of outputs on certain concepts
\cite{kim_interpretability_2018}.
The vector can be the normal vector of a linear model, or the center
of a cluster, and is called the embedding of the
concept~\cite{fong_net2vec_2018}.

While concept analysis opens up a wide variety of verification
options, existing methods suffer from some limitations.
For one, the methods are either restricted to image-level concepts,
or, like Net2Vec~\cite{fong_net2vec_2018}, report poor results for
part objects.
But part objects, such as human body parts, are substantial to verify
logical plausibility of pedestrian detectors
needed for automated driving. 
This brings up the second issue: To our knowledge, existing proposals
have not been evaluated on large networks like standard object
detectors.
Concretely, we uncovered some severe performance limitations of
Net2Vec that impede application to state-of-the-art sized DNNs or
larger concept datasets than the original Broden dataset proposed
in~\cite{bau_network_2017}.

In order to overcome the practical issues, this paper suggests some
substantial improvements to the Net2Vec method, demonstrated on a
new large MS~COCO~\cite{lin_microsoft_2014} based concept dataset for
human body parts.
The applicability of concept analysis for simple verification is
demonstrated by exemplary assessment of three networks with
different sizes, tasks, and training data (AlexNet, VGG16, and
Mask~R-CNN~\cite{he_mask_2017}).
Concretely, the semantic similarity of different concepts is
validated, and it is checked whether the internal representation of
the networks is biased towards one size category (\idest distance to the
camera).
Interestingly, our findings suggest that convolutional DNNs are not
biased towards one size. Instead, from early layers onwards, they use an
efficient common representation for instances of the same part object
but different sizes.
%
%
%
%
%
The main contributions of this paper are:
\begin{itemize}[nosep]
\item A Net2Vec-based concept analysis approach 
  that can deal with large models and large concept datasets
  is presented and demonstrated.
  It is found to considerably surpass the state-of-the-art for object
  part concepts.
\item The first evaluation of size invariance of the
  internal representations of convolutional DNNs is conducted.
\item For this, an approach is presented to estimate the
  pixel size of persons in 2D images from skeletal annotations,
  which can be used for concept mask generation.
\end{itemize}
After a revision of concept analysis methods in
\autoref{sec:relatedWork}, our approach is introduced in
\autoref{sec:approach}. Finally, in \autoref{sec:experiments} it is
validated and used for the exemplary assessment.

\section{Related Work}\label{sec:relatedWork} 

An early approach to supervised concept analysis was
NetDissect by Bau et al.~\cite{bau_network_2017}. They associated single
convolutional filters with semantic concepts, and provided the Broden
dataset, a combined dataset featuring a wide variety of concepts.
%
Building upon NetDissect, Net2Vec~\cite{fong_net2vec_2018} expanded
from single filters to combinations of filters and trains a
$1\times1$ convolution to predict concept masks from latent space
outputs.
Intuitively, the weight vector of the convolution, the concept
embedding vector, encodes the direction within the layer output space that
adds the concept. The authors investigate the similarity of different
concepts via cosine similarity of their concept vectors.
%
Similarly, TCAV by Kim et al.~\cite{kim_interpretability_2018} trains
linear models but using a support vector machine on the complete latent
space output for binary classification instead of segmentation.
The authors suggest to use partial directional derivatives
along the concept vectors
to assess how outputs depend on concepts in intermediate layers.
Future work may investigate how this can be applied to concept
segmentation instead of concept classification.
%
%
Other than the previous linear model approaches,
SeVec~\cite{gu_semantics_2019} uses a k-means clustering approach,
demonstrated on a dataset other than Broden, but again only on
image-level concepts and with a manual step in the process.
The Net2Vec-based proof of concept in \cite{schwalbe_concept_2020}
uses a small concept dataset for traffic sign letters.
This work also details the value of concept analysis for
safety assessment, and suggests that a mismatch of concept size and
receptive field of concept models can cause a texture
bias, that is, predictions purely based on texture.
This is investigated for larger nets in this paper.
%

More relevant for inspection than requirements verification purposes,
unsupervised concept analysis approaches aim to
find and visualize repetitive concepts in the intermediate output
of a DNN. 
%
A simple example is feature visualization~\cite{olah_feature_2017},
where noise is optimized to maximally activate a convolutional filter.
%
ACE~\cite{ghorbani_automatic_2019}
instead uses super-pixels as concept candidates, which are then
clustered by their latent space proximity.
%
In the direction of representation disentanglement,
\cite{yeh_completenessaware_2020} introduced a measure for the
completeness of a set of concept vectors with respect a given
task. Their idea is that post-hoc adding a semantic bottleneck
with unit vectors aligned to the concepts should not decrease
the model performance.
Similarly, but with invertible DNNs instead of linear maps,
the recent work in \cite{esser_disentangling_2020} tries to
find a bijection of a latent space to a product of semantically
aligned sub-spaces (and a residual space). 

Experiments so far have been conducted on small resolution image datasets
like the $224\times224$ pixels of the Broden
\cite{bau_network_2017,fong_net2vec_2018,kim_interpretability_2018}
and ImageNet \cite{gu_semantics_2019} data or
less \cite{schwalbe_concept_2020}, and have not yet been applied
to large models or object detectors.


\section{Approach}\label{sec:approach} 
This section introduces our concept analysis approach with its
modifications to the Net2Vec concept analysis method,
and an approach to estimate the body size in
pixels for persons in 2D images from skeletal annotations.
Size estimation is later used to generate ground truth segmentation
masks for body part concepts, and to categorize them by size.

\subsection{Concept Embedding Analysis} 

\begin{figure}
  \centering
  \strut\\[-1.5\baselineskip]
  \hspace*{-0.5em}%
  \parbox[t]{0.5\linewidth}{\strut\\[-2.5\baselineskip]\raggedleft
    \includegraphics[height=4.5cm]{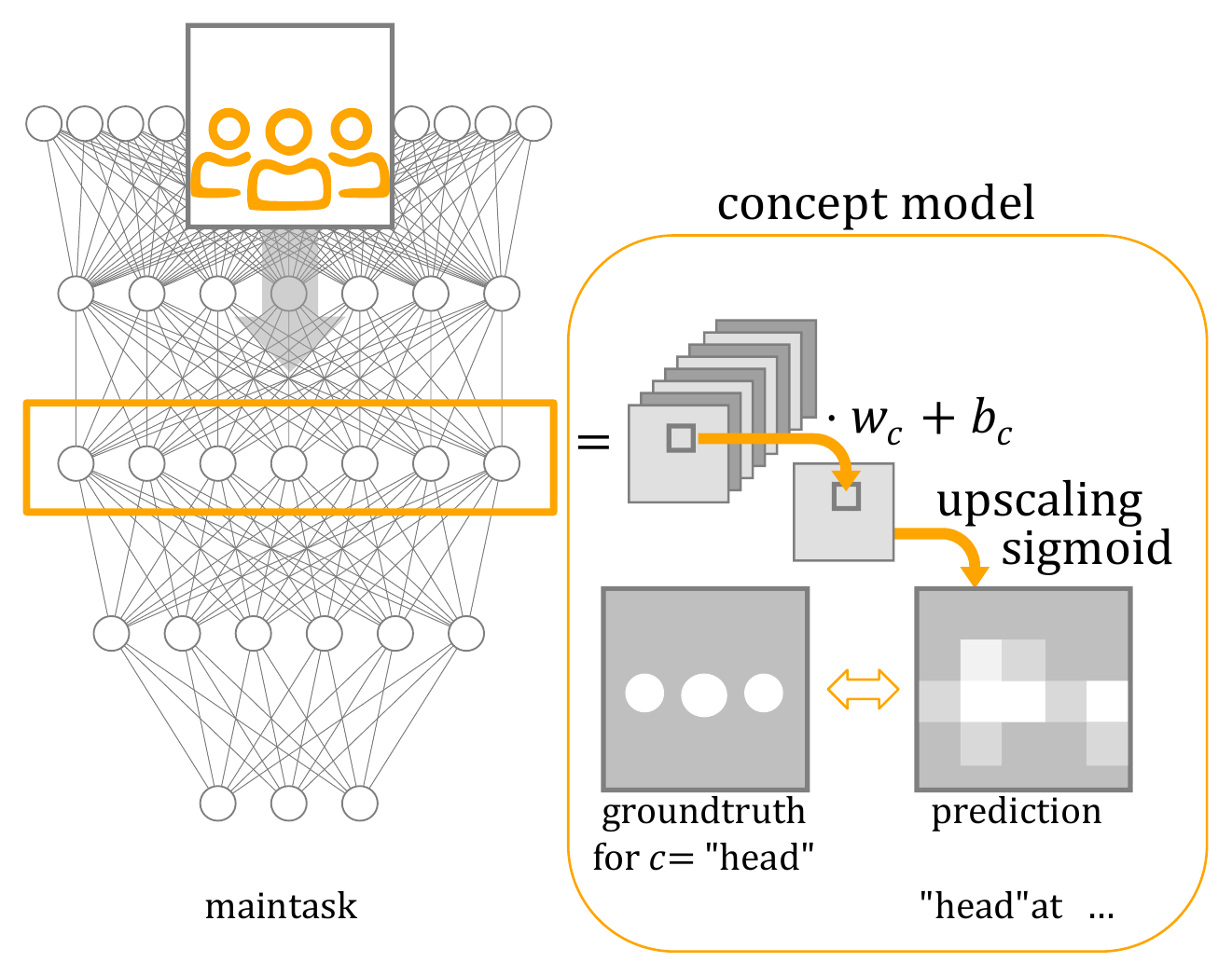}
  }\hspace*{1em}%
  \parbox[t]{0.5\linewidth}{
    \caption{Illustration of the used concept analysis approach.
      For a concept $c$, here \enquote{head}, it predicts binary
      segmentation masks from convolutional activation maps via
      (1)~a 1-filter convolution with kernel $w_c$ and bias $b_c$,
      (2)~bilinear upscaling
      (here depicted using nearest pixel upscaling),
      and finally (3)~normalization.}
    \label{fig:conceptAnalysis}
  }
\end{figure}


Our method is illustrated in \autoref{fig:conceptAnalysis}.
As a foundation, the Net2Vec \cite{fong_net2vec_2018} approach was chosen.
It allows to localize non-image-level concepts like object parts in
convolutional activation maps by predicting binary segmentation masks.
For this the spatial resolution of the convolutional layers is used:
The mask predictor is a $1\times1$ convolution followed by bilinear
upscaling, then sigmoid normalization (and, for binarizing, thresholding at 0.5).
For comparability, the quality measure set intersection over union
(set IoU) is adopted
which divides the total area of intersections by the total area of unions
between the binary ground truth masks $M_i$ and the predicted
segmentation masks $M_i^\text{pr}$ (binarized at 0.5):
\begin{gather}
  \SwapAboveDisplaySkip
  \text{set IoU}\left((M_i)_i, (M_i^{\text{pr}})_i\right)
  = \frac{\sum_i \sum M_i \cap (M_i^\text{pr}>0.5)}{\sum_i \sum M_i \cup (M_i^\text{pr}>0.5)}
  \;.
\end{gather}
We here took the mean of batch-wise set IoU values.
It must be noted that the set IoU measure penalizes errors on small
objects more than on large ones, and it suffers under low resolution
of activation maps.
Below, four further limitations of Net2Vec are collected and our
countermeasures explained.

Most notably, Net2Vec uses denoising of activation maps with a
ReLU suggested in \cite{bau_network_2017}.
The threshold is calculated to keep the 0.5\% highest activations.
This requires to load activations of the complete dataset into memory
at once, which is infeasible both for large models and large datasets.
Experiments showed that denoising makes no noticeable
difference, hence it is skipped.

Next, to push the decision boundary of the model towards a value
of 0.5, Net2Vec uses a binary cross-entropy (BCE) loss with per-class-weights
where each weight is the mean proportion of pixels of the opposite
class.
Obtaining the weight is an expensive pre-processing step.
Here we found that global weighting could be replaced either by
calculating weights batch-wise, or using an unweighted
Tversky loss~\cite{salehi_tversky_2017}---also called Dice or F1
loss---that directly optimizes the F1 score of the segmentation
per $h\times w$ image (\enquote{$\cdot$} pixel-wise):
\begin{gather}
  \text{Dice loss}\left(M, M^{\text{pr}}\right)
  = 1 - \frac{2 \sum M \cdot M^\text{pr}}
  {\sum M + \sum M^\text{pr}}
  \quad \text{for $M,M^\text{pr}\in[0,1]^{h\times w}$.}
\end{gather}

Further, the original method reports average set IoU values lower
than 0.05 for part objects, which we could confirm.
This can be achieved by constantly predicting white masks
(1 for all pixels).
In a series of experiments we found settings fixing this:
We use a Tversky loss, Adam optimizer \cite{kingma_adam_2015}
instead of stochastic gradient descent (SGD),
with a batch size of 8 instead of 64,
and learning rate of \num{e-3} rather than \num{e-4}.

Another modification used in some experiments is
\emph{adaptive kernel size}.
The original Net2Vec method uses a $1\times1$ kernel, so each
prediction only has a spatial context of one pixel. As found in
\cite{schwalbe_concept_2020}, this may be too small for larger
concepts and layers with higher resolution, and possibly leads to a
texture bias~\cite{geirhos_imagenettrained_2019}.
Following \cite{schwalbe_concept_2020}, this can be mitigated using
a kernel size that covers an area of the expected size of the
concept, assuming low size variance.
Note that this may improve performance of a concept model,
but destroys comparability amongst models of differently sized
concepts due to incompatible kernel shapes.

\subsection{Distance Category Estimation} 
\label{sec:sizeEstimation}

\begin{figure}
  \strut\\[-2\baselineskip]
  \scriptsize
  \centering
  \hspace*{\fill}%
  \begin{tabulary}{0.2\linewidth}[t]{@{}r@{\,$\cdot$\,}>{\ttfamily}l @{}}
    \toprule%
    \multicolumn{2}{c}{Formulas for total height}\\\midrule[\heavyrulewidth]
    1    & bbox\_width/\_height \\\midrule 
    1.1  & body\_height         \\\midrule 
    2.4  & hip\_to\_shoulder    \\\midrule 
    7    & head\_height         \\\midrule 
    1.485& leg $\cdot \frac{h}{h-0.433}$        \\\midrule 
    2.77 & upper\_leg  $\cdot \frac{h}{h-0.405}$ \\\midrule 
    3.075& lower\_leg $\cdot \frac{h}{h-0.501}$ \\\midrule 
    3.72 & upper\_arm $\cdot \frac{h}{h-0.449}$ \\\midrule 
    4.46 & lower\_arm $\cdot \frac{h}{h-0.569}$ \\ 
    \bottomrule
  \end{tabulary}%
  \hspace*{\fill}%
  \begin{minipage}[t]{0.17\linewidth}
    \strut\\
    \includegraphics[width=\linewidth]{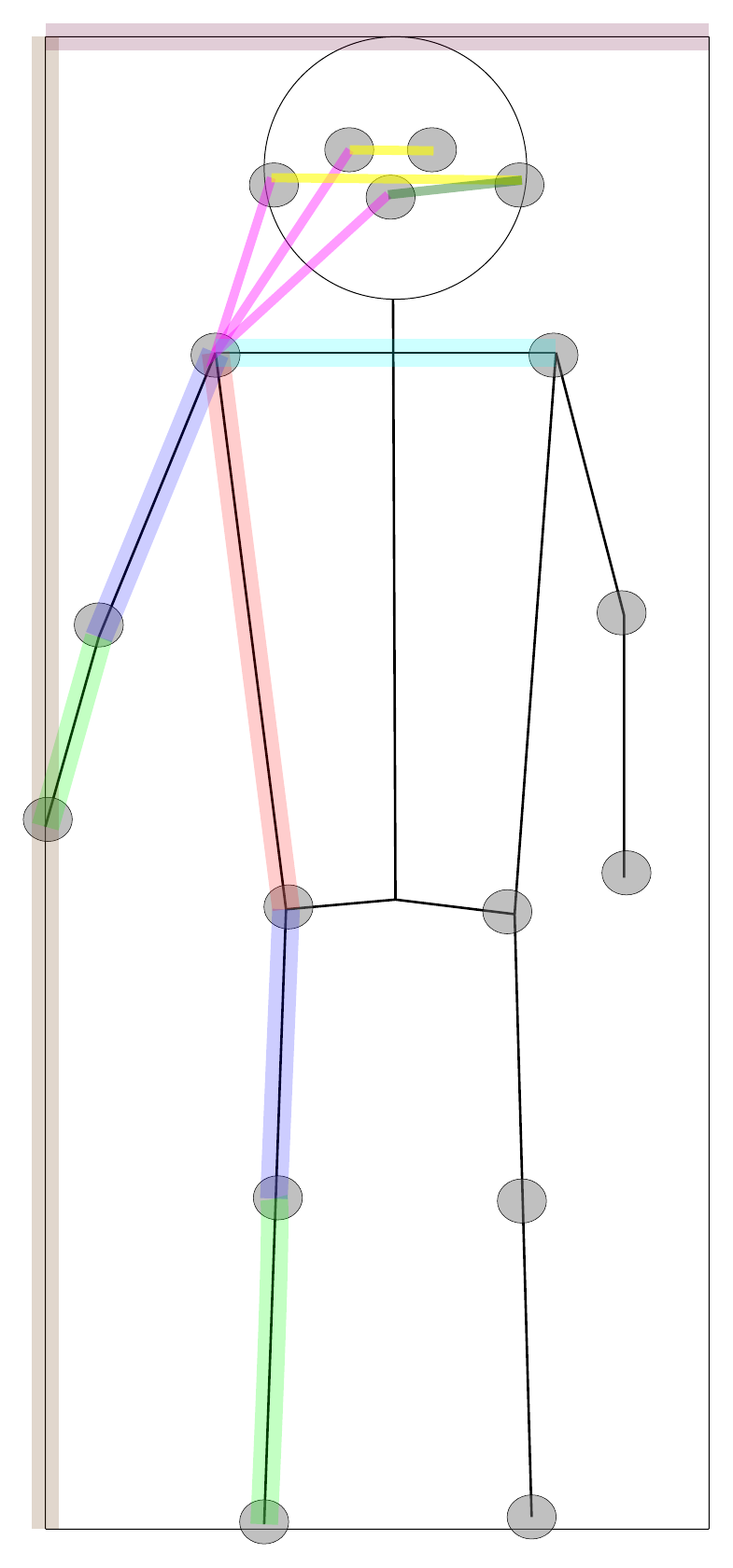}
  \end{minipage}%
  \hspace*{\fill}%
  \begin{minipage}[t]{0.45\linewidth}
    \begin{tabulary}{\linewidth}[t]
      {@{}>{\ttfamily}l@{~$\approx$~} >{\ttfamily}L@{}}
      \toprule%
      \multicolumn{2}{l}{Formulas for derived lengths} \\
      \midrule[\heavyrulewidth]
      body\_height
      & leg\textsuperscript{*} + hip\_to\_shoulder\textsuperscript{*} + shoulder\_to\_eye\textsuperscript{*}/\_ear\textsuperscript{*}/\_nose 
      \\ & arm + shoulder\_width
      \\\midrule 
      leg
      & lower\_leg + upper\_leg
      \\\midrule 
      arm
      & lower\_arm + upper\_arm
      \\\midrule
      head\_height
      & $1.1\,\cdot$\,head\_width
      $\approx$~$\frac{8}{7}$\,head\_depth
      \\\midrule
      head\_width
      & ear\_to\_opposite\_ear
      \\ &$2.5\,\cdot$\,eye\_to\_eye \\\midrule 
      head\_depth
      & $2\,\cdot$\,ear\_to\_eye\textsuperscript{*}
      $\approx$~$\frac{7}{4}$\,ear\_to\_nose
      \\ 
      \bottomrule
    \end{tabulary}
    \\{\tiny\textsuperscript{*}\,involved keypoints must be on the same side}
  \end{minipage}%
  \hspace*{\fill}
  \caption{
    Used keypoints and links with formulas to estimate the total
    body height of a person from link lengths and
    assumed \enquote{standard} height $h$ in meters.
  }
  \label{fig:sizeFormulas}
\end{figure}


Assume an image dataset is given that provides 2D skeletal information
of persons (cf.~\autoref{fig:sizeFormulas}), possibly bounding boxes,
but no depth information.
The goal of this approach is to estimate and categorize the body size
of a person depicted in a 2D image in pixels, given only lengths
between keypoints.
The challenges are that keypoint information may be incomplete due to
occlusion or cropping, and that 2D projected lengths may be
inaccurately short.
Thus, redundant calculation of body size from as many link types as
possible is needed.
One major drawback of this method is that assumptions on
\enquote{standard} proportions must be made.
This cannot be overcome without additional information.
Despite that, it was found that the approach yields surprisingly good
and intuitive results, with errors not infringing the quality of body
size categorization.

The following sources were used to relate the true length $l$ of
(combinations of) keypoint links to the true body height $h$:
arts 
for facial and head-to-body proportions~\cite{larson_standard_2014};
standard educational material relating the wrist-to-wrist span
\cite{debrabandere_human_2017}; and
linear models used in archeology to estimate the body size from
single long bones~\cite{larson_standard_2014}.
For long bone relations, the mean model parameters were taken
between genders. 
Relations involving approximate \texttt{body\_height} and bounding box
(\texttt{bbox}) dimensions were estimated manually.
The resulting relation models are all of the linear nature
$h = s\cdot l + c$ for some slope $s$, and offset $c$ in meters.
%
Given the downscaled length $l'=f\cdot l$ in pixels, the formula for
the downscaled body size is
$f\cdot h = l' \cdot \frac{h}{h - c}$.
If $c$ is non-zero, a \enquote{standard} real person size $h$ must be
assumed, which was set to \SI{1.7}{\metre} following
\cite{larson_standard_2014}.
The final formulas are shown in \autoref{fig:sizeFormulas}.
Whenever one value has several estimation results,
the maximum was used to cope with possible length
shortenings due to 2D projection.

The formulas were now leveraged to sort annotations into four size
categories. These were chosen such that in each the minimum and
maximum estimated size do not differ by more than a factor of two.
Very small and very large persons were discarded.
The resulting size categories with their range of relative sizes are:
\emph{far} in $[0.2, 0.38]$,
\emph{middle} in $[0.38, 0.71]$,
\emph{close} in $[0.71, 1.33]$, and
\emph{very close} in $[1.33, 2.5]$.
Categories are depicted in \autoref{fig:cocoSizeDistribution}.

\section{Experiments}\label{sec:experiments}
After detailing our experiment settings, this section will
validate the used approach including hyperparameters
(\autoref{sec:validationApproach}),
then conduct exemplary assessments of embedding quality and
semantics (\autoref{sec:embeddingValidation}),
as well as size bias (\autoref{sec:sizeBiasValidation}).

\begin{figure}
  \centering
  \includegraphics[width=\linewidth]{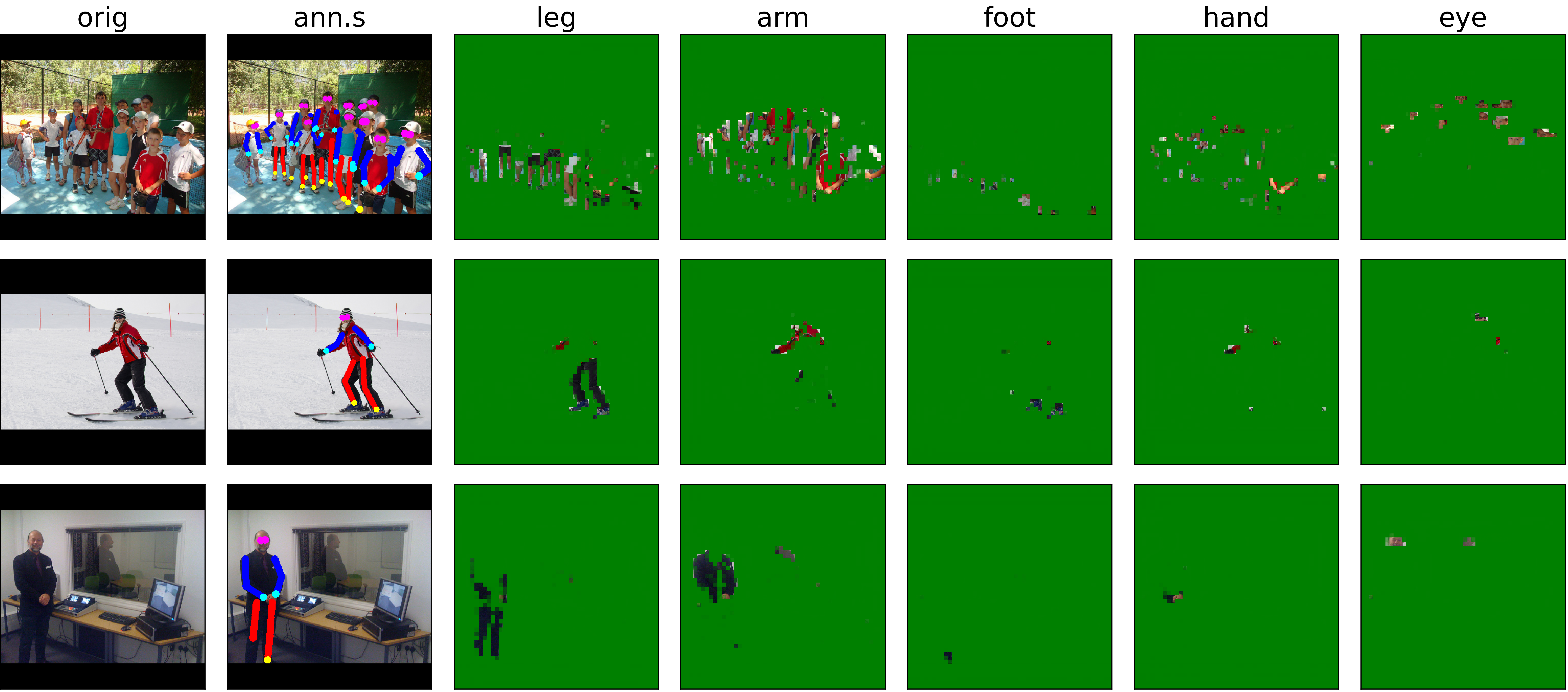}
  \caption{
    Overlay (green) of segmentation masks predicted by the mean
    concept model of Mask~R-CNN backbone layer block~3.
    Original images are shown in first column, generated ground truth
    annotations in second column in different colors.
    Mean is taken over normalized concept vectors,
    cf.~\cite{rabold_expressive_2020}.
    Masks are upsampled using nearest rule to demonstrate resolution.
    Note how predicted masks clearly correlate with the ground truth
    instead of being purely white.
    Images are taken from test set (MS~COCO \texttt{val2017}).
    {\tiny
    For the used images thanks to:
      top: Oleg Klementiev
      \url{http://farm5.staticflickr.com/4115/4906536419_6113bd7de4_z.jpg}
      \textcopyright~\href{http://creativecommons.org/licenses/by/2.0/}{CC~BY~2.0};
      middle: Nick Webb
      \url{http://farm8.staticflickr.com/7015/6795644157_f019453ae7_z.jpg}
      \textcopyright~\href{http://creativecommons.org/licenses/by/2.0/}{CC~BY~2.0};
      bottom: Yandle
      \url{http://farm4.staticflickr.com/3179/2986591710_d76622fdf0_z.jpg}
      \textcopyright~\href{http://creativecommons.org/licenses/by/2.0/}{CC~BY~2.0}
    }
  }
  \label{fig:sampleOutputs}
\end{figure}

For the experiments, we chose three networks of different
size, training objective, and training dataset, with pretrained
weights from the 
\href{https://pytorch.org/vision/stable/models.html}
{torchvision model zoo}\footnote{
\url{https://pytorch.org/vision/stable/models.html}}:
ImageNet-trained classifiers
VGG16~\cite{simonyan_very_2015} and
AlexNet~\cite{krizhevsky_one_2014},
and the object detector Mask R-CNN~\cite{he_mask_2017}
trained on MS~COCO. 
%
As layers we chose the activated convolutional output before each
downsampling step except the first.
In case of Mask R-CNN we considered the output of each residual
grouping in the backbone and the feature pyramid.
%
For concepts we selected a set of five large and small
concepts that are common to the Broden and MS~COCO datasets:
\concept{leg}, \concept{arm}, \concept{foot} and \concept{hand}
(for COCO approximated via ankle and wrist keypoints), and
\concept{eye}.
%
Images were zero-padded to square size then resized to
$400\times400$ respectively $224\times224$ (VGG16, AlexNet) pixels to
avoid feature distortion.
Ground truth masks on concepts for MS~COCO images were generated by
drawing links (\concept{leg}, \concept{arm}) respectively points
(\concept{foot}, \concept{hand}, \concept{eye}) of
width 0.025 times the image height or times the body size if this can
be estimated (see \autoref{fig:sampleOutputs}).
Intermediate outputs of the DNNs were cached with bi-float~16
precision to speed up experiments, except for the very large
Mask R-CNN feature pyramid blocks~0 and~1, and backbone layer~1.
On cached layers, concept model training was done with 5-fold
cross-validation.
The original train-test splits were used.

\subsection{Validation of the Proposed Methods}\label{sec:validationApproach}

\begin{figure}
  \centering
  \mbox{
    \hspace*{-0.05\linewidth}
    \begin{minipage}[t]{0.35\linewidth} 
      \strut\\[-\baselineskip]
      \includegraphics[width=1.07\linewidth]{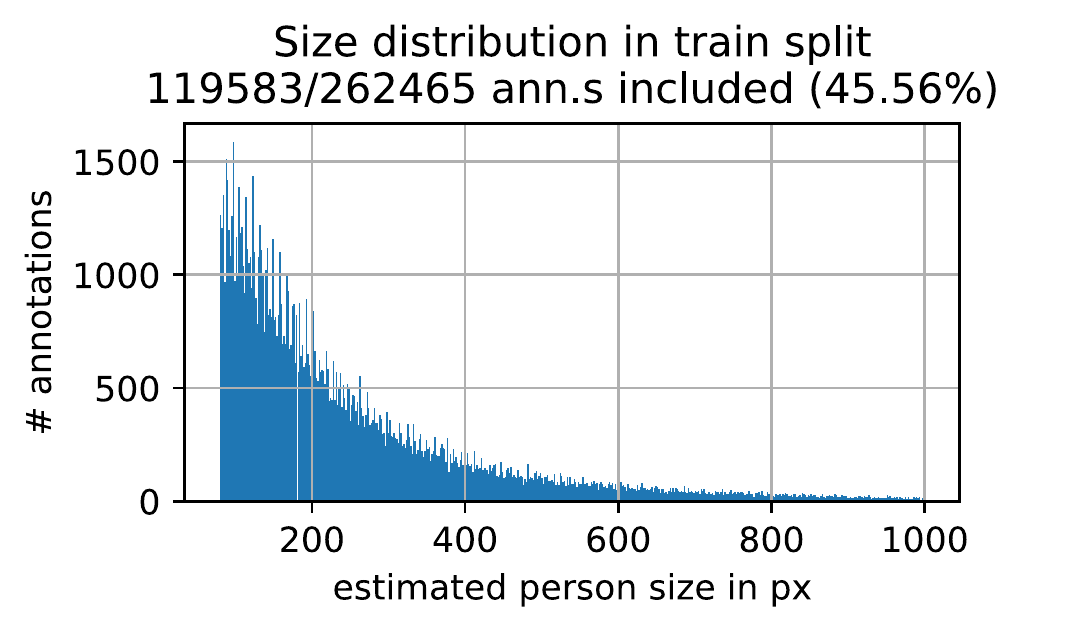}
    \end{minipage}
    \begin{minipage}[t]{.43\linewidth} 
      \strut\\[-\baselineskip]
      \includegraphics[width=1.05\linewidth]{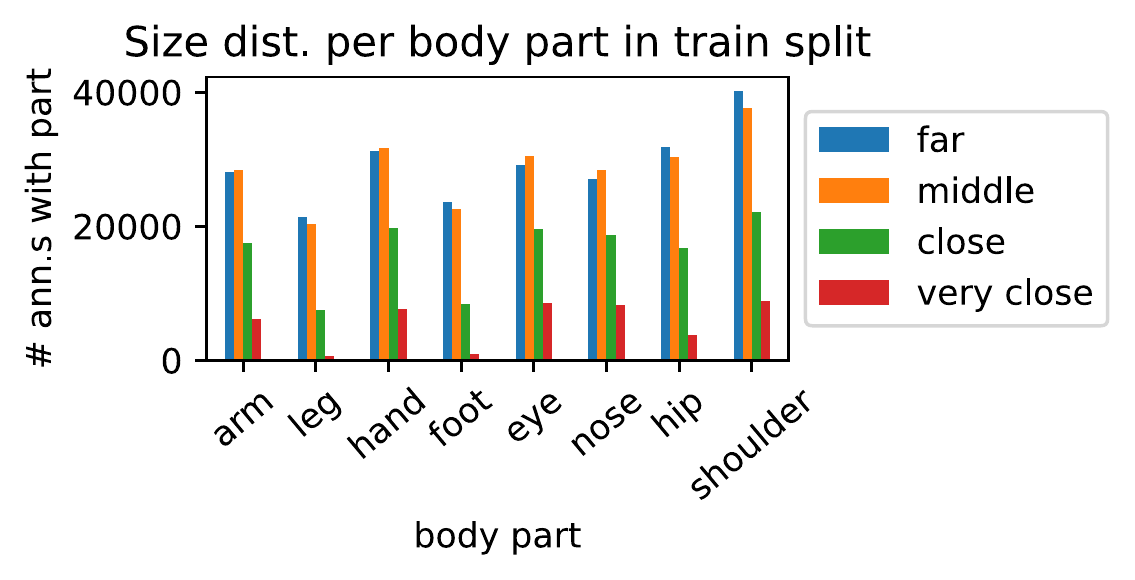}
    \end{minipage}
    \parbox[t]{0.25\linewidth}{\strut\\[-2\baselineskip]
      \includegraphics[width=\linewidth]{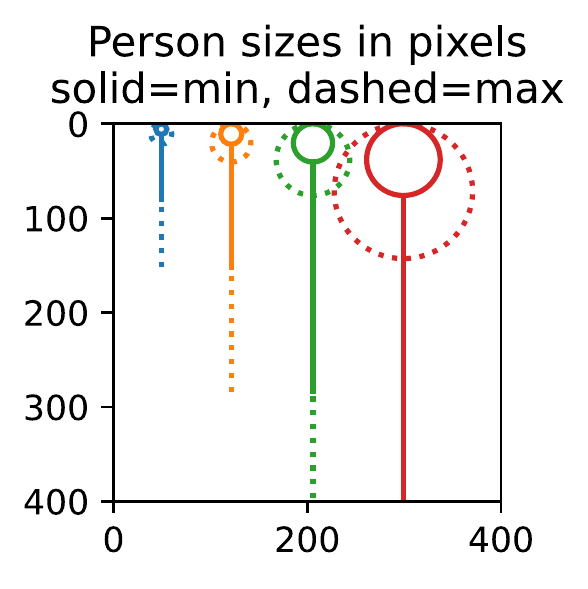}
    }
  }\\%
  \hspace*{-0.05\linewidth}%
  \includegraphics[width=1.1\linewidth]{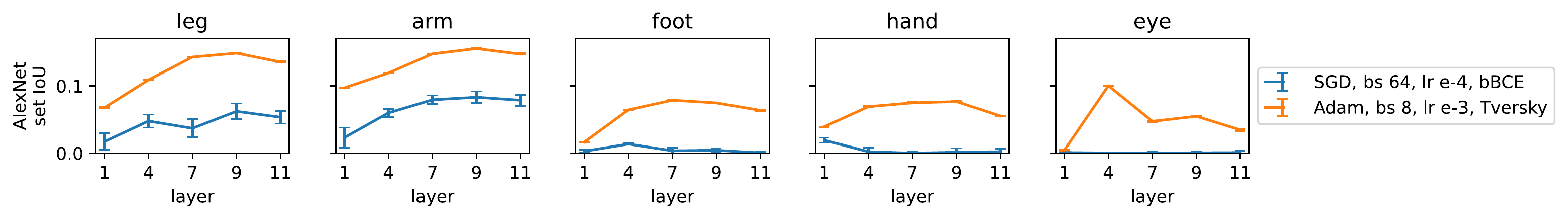}
  \caption{
    \emph{Top:}
    Distribution by estimated person size of all annotations (\emph{left})
    and per body part (\emph{center}) for COCO train images
    (padded to square size and resized to $400\times400$\,px);
    used person size categories are illustrated on the \emph{right}.
    \emph{Bottom:}
    Performance comparison of used settings with Net2Vec baseline
    (cf.~\cite[Fig.~2]{fong_net2vec_2018}).
    Best viewed in color.
  }
  \label{fig:cocoSizeDistribution}
\end{figure}

\paragraph{Size Distribution in MS~COCO}
To validate the size estimation approach, an analysis of the size
distribution on the popular MS~COCO dataset was conducted.
The size of an annotation cannot be estimated if it provides
no keypoints (very small persons and crowds) or only disconnected
keypoints. Nevertheless, size estimations could be made for a decent
proportion of the annotations in both training and test set (ca. 46\%),
see \autoref{fig:cocoSizeDistribution}.
Also, for each chosen body part, a sufficient amount of annotations is
available in size categories \emph{far}, \emph{middle}, and \emph{close}:
more than 10,000 in the train, and 300 in the test set.
These categories were thus used for experiments.
Training and test set were found to be similarly distributed,
with significantly more keypoint annotations estimated far than
close. This suggests, that the total amount of positive pixels for
masks of small far body parts is similar to that of large close ones.

\paragraph{Improvement of Net2Vec}
Settings for the concept analysis approach were needed that
(1) work for large models and datasets (no expensive thresholds
and weights), and
(2) work for part objects.
First, the necessity to threshold the activation maps was disproved:
Skipping it had no effect on the test performance.
Then, without thresholding, a series of experiments was conducted
comparing different
optimizers (SGD, Adam),
batch sizes (bs), learning rates (lr), and
losses (globally weighted BCE, batch-wise weighted BCE, and Tversky).
Each setting was evaluated on AlexNet and VGG16, on all selected
layers and concepts, using the original Broden
dataset~\cite{bau_network_2017}.
Firstly, the better optimizer was chosen, then the better batch-size
and learning rate combination, and lastly losses were compared.
The consistently best and stable setting proved to be Adam optimizer
with lr~\num{e-3}, bs~8, with maximum five epochs due to fast convergence.
Find a comparison to the baseline for AlexNet in
\autoref{fig:cocoSizeDistribution} (VGG16 similar).


\subsection{Embedding Quality and Similarity Validation}
\label{sec:embeddingValidation}

Two applications of concept analysis are to verify sufficient
embedding qualities, and to use the similarity
measure on the concept models for validation of semantic relations.
For this, concept models for all chosen concepts, networks, and layers
were trained on the complete MS~COCO concept dataset.
Performance results are depicted in \autoref{fig:sizeComparison}
(size category \emph{all}).
%
As before, the variance was very low, evidencing good convergence
of the linear concept models.
%
Convincing embedding qualities were found except for the small
concepts \concept{hand} and \concept{foot} in small networks.
This may origin from the approximate and noisy ground truth, and the
higher set IoU penalties for low resolution.
%
In general, later layers with low resolution showed the best embedding
results, indicating that body parts are relatively complex concepts.
%
These may require larger network structures, since Mask R-CNN
significantly surpassed AlexNet and VGG16.
Some exemplary output for Mask~R-CNN is shown in
\autoref{fig:sizeComparison}.

\begin{figure}
  \hspace*{-0.07\linewidth}%
  \parbox[t]{0.63\linewidth}{\strut\\[-1.1\baselineskip]
    \includegraphics[height=2.1cm]{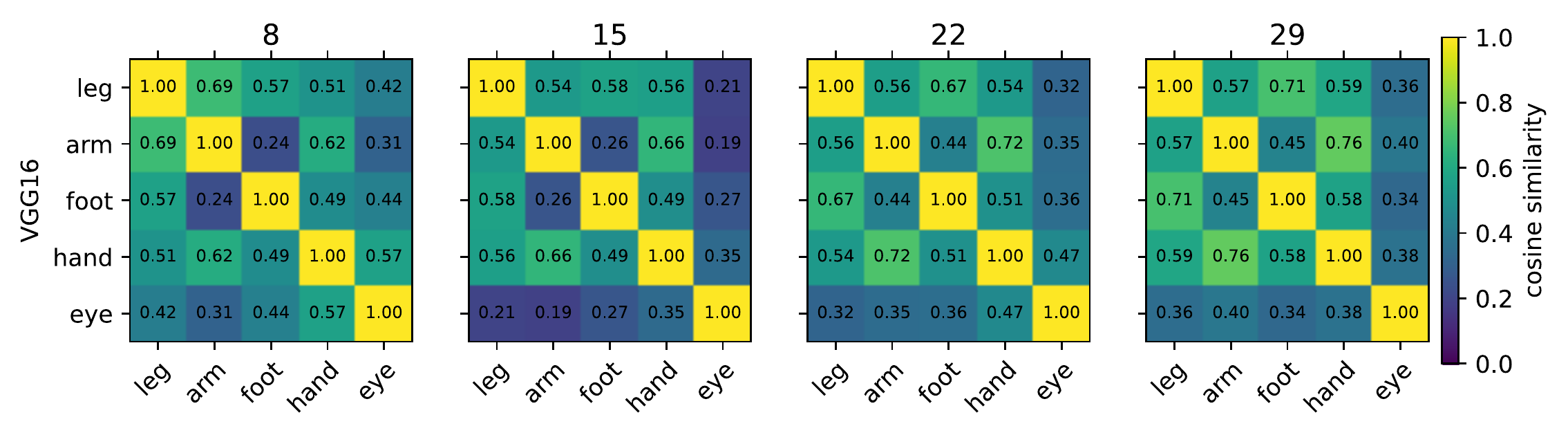}%
  }%
  \begin{minipage}[t]{0.5\linewidth}\strut\\[-2.1\baselineskip]
    \caption{
      Mean cosine similarities between normal vectors of concept
      models for different body parts, by layer.
      AlexNet results were similar to VGG16 (\emph{top}).
    }\label{fig:cosineSimilarities}
  \end{minipage}\\[-.5em]
  \hspace*{-0.07\linewidth}%
  \includegraphics[height=2.1cm]{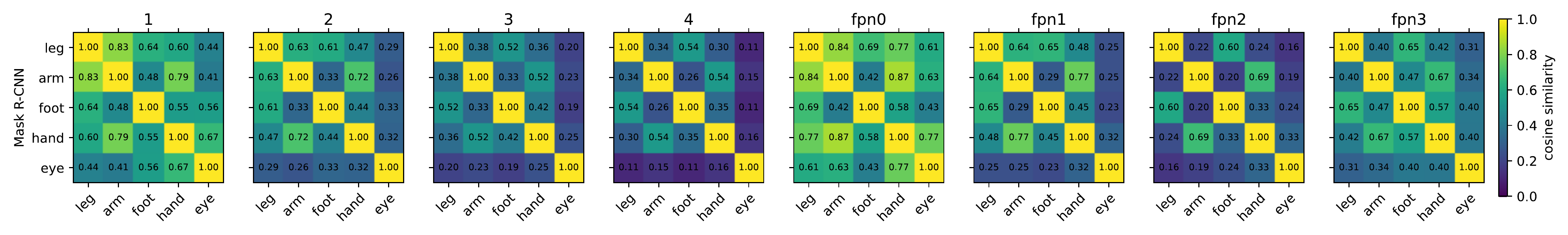}%
  \\[-.5em]
\end{figure}

Next, the concept model similarities were measured as the mean cosine
similarity between the normal vectors of the models, shown in
\autoref{fig:cosineSimilarities}.
Cosine similarity $\frac{a\cdot b}{\|a\|\cdot\|b\|}$ of vectors
$a,b\in \mathds{R}^n$
is the cosine of the angle between the vectors, with 1 meaning 0°
angle, and -1 meaning 180°.
The results show intuitive relations between the concepts:
%
No similarity values were below zero, so no body part concepts are mutually exclusive.
%
And \concept{eye} is most different from the other parts,
%
while \concept{hand} and \concept{arm} are more closely related,
similarly \concept{leg} and \concept{foot}.
%
Interestingly, in the later low-resolution layers
concepts were more dissimilar, so better distinguishable,
explaining the better performance here.

\subsection{Correlations of Person Size and Segmentation Quality}
\label{sec:sizeBiasValidation}
The last series of experiments means to investigate the
following questions:
\begin{enumerate*}[label=\arabic*), nosep]
\item \label{claim:sizeDifference}
  Does the internal representation of a concept differ for different
  sizes of the concept?
  And is there a safety critical bias towards one size
  (\idest distance from the camera)?
\item \label{claim:adaptiveNeeded}
  Is adaptive kernel size needed to avoid texture bias?
\end{enumerate*}
A yes to \ref{claim:sizeDifference} would suggest that different and
better embeddings can be found if concept information is assessed
separately for each size category.
Then, especially for larger size categories, adaptive kernel sizes
might be necessary.
%
To answer the questions, results on a test set restricted to one
size category were collected on the MS~COCO concept data.
This was done and compared for concept models trained on the complete
dataset (all nets), and models trained on the respective size category
either with $1\times1$ or adaptive kernel size (AlexNet, VGG16).
%
Adaptive kernel sizes were chosen to
cover the following areas relative to the mean person size of the
size category (in height$\times$width):
$0.3\times0.1$ for \concept{leg},
$0.2\times0.15$ for \concept{arm},
$0.1\times0.1$ for \concept{foot} and \concept{hand}, and
$0.04\times0.04$ for \concept{eye}.
The two main results were: The internal representation of body parts
seems to be mostly size invariant, and adaptive kernel sizes are not
necessary. 

\begin{figure}
  \includegraphics[width=\linewidth]{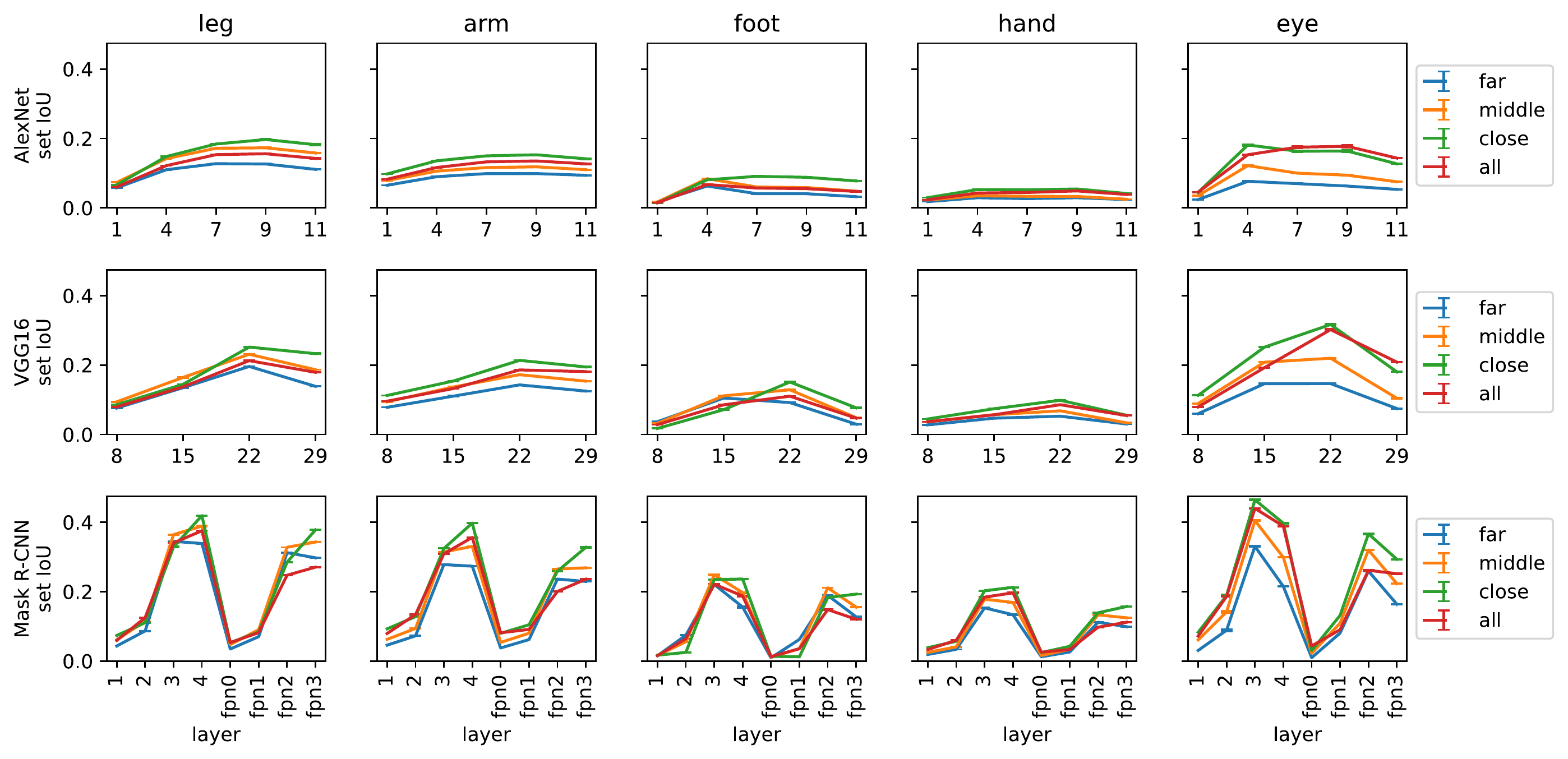}
  \caption{
    Test set IoU results of concept models trained on
    the complete dataset, by layer, annotation size category,
    model (top to bottom), and concept (left to right).
    Note that standard deviations are marked but negligible.
    Best viewed in color.
  }\label{fig:sizeComparison}
  ~\\
  \hspace*{-0.05\linewidth}%
  \includegraphics[width=1.1\linewidth]{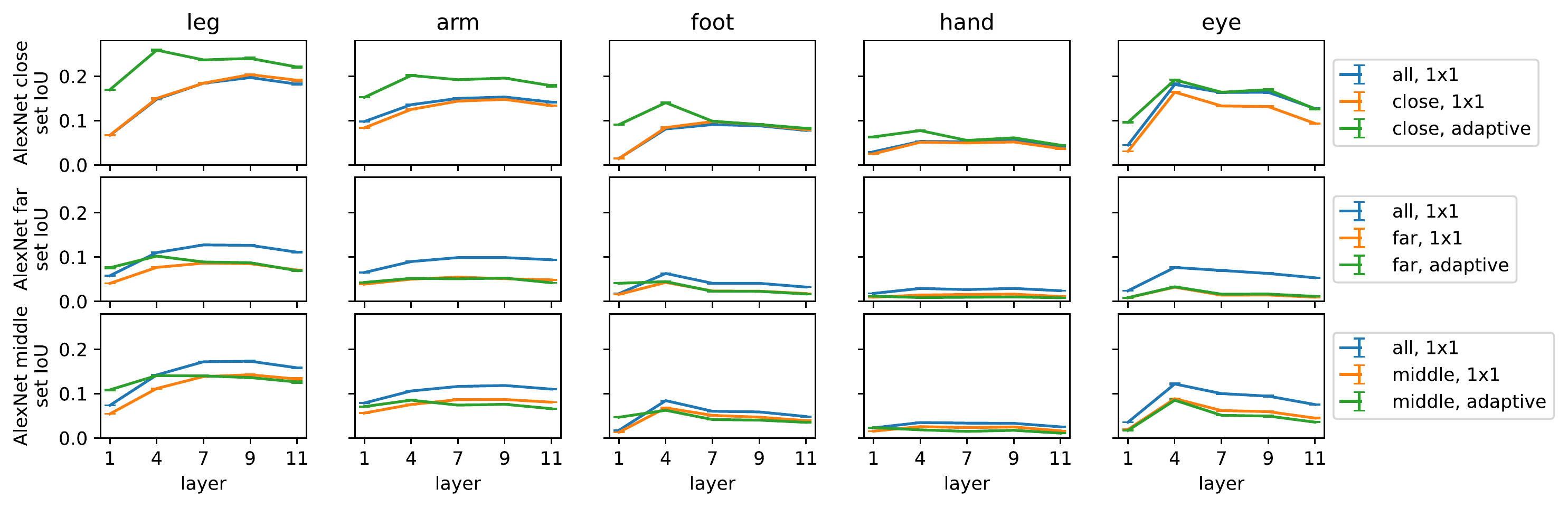}\\%
  \hspace*{-0.05\linewidth}%
  \includegraphics[width=1.1\linewidth]{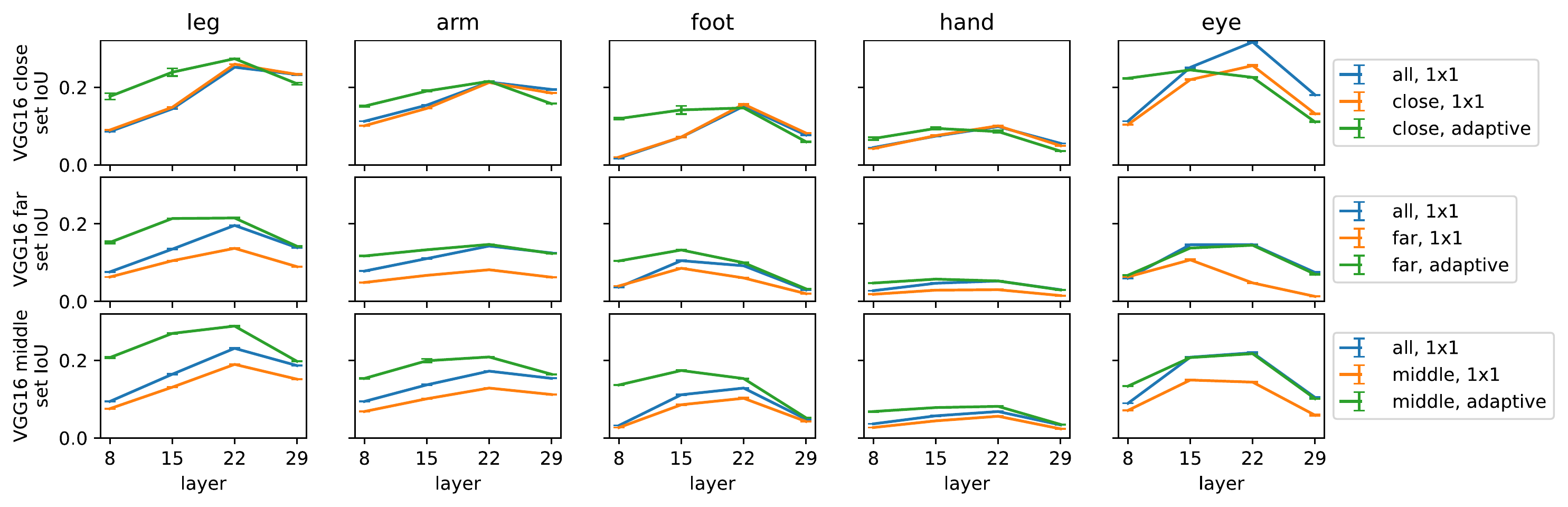}
  \caption{
    Test set IoU results of concept models trained on a size subset
    by layer and kernel setting.
    Results were obtained on the test set restricted to the respective
    training size category. Results for the models trained on all
    data are added for comparison, cf.~\autoref{fig:sizeComparison}.
    Best viewed in color.
  }\label{fig:kernelComparison}
\end{figure}

\paragraph{Results}
In \autoref{fig:sizeComparison} the performance of concept models
trained on \emph{all} data was compared for different test subsets.
The differences between size categories did not exceed what was
expected from the different set IoU penalties, so no bias could be
found here.
Since a concept model trained on all data may not be the optimal one
for a size, we also compared the \emph{all} models to models
solely trained on the respective size category, see
\autoref{fig:kernelComparison}.
This showed that a restricted training set may even decrease test
results, indicating that at least parts of the internal representation
used for the different sizes must be shared.
To substantiate that, the normal vectors of the linear models
for the different size categories were compared.
In the best performing layers the restricted size
categories had high cosine similarity (over 0.7) to \emph{all},
with \emph{far} deviating the most.
However, \emph{all} could not be represented as a linear
combination of the sub-sizes, and cosine similarities of least squares
solutions did not top 0.95.
So, some information even seems to get lost when training only on single
size categories.
Lastly, results on single size categories were also compared to
adaptive kernel size results in \autoref{fig:kernelComparison}.
Adaptive kernel size would show some improvements
but the best layer rarely changed.

\section{Conclusion} 
This paper proposed an efficient concept analysis approach
that is fit for practical application to large networks and datasets,
and to object part concepts.
It was demonstrated how this method can be used to assess consistency
and size bias in the internal representations of a DNN.
For this, a size estimation method from 2D skeletal data is
proposed together with a new concept dataset based on MS~COCO keypoint
annotations.
Our assessment results suggest that the representations within
standard models AlexNet, VGG16, and Mask R-CNN are mostly invariant to
different sizes respectively camera distances.
%
Despite drawbacks of the used set IoU metric,
%
concept analysis is shown to be a promising approach for verification
and validation of deep neural networks.
Future work will investigate further assessment possibilities arising
from post-hoc explainable access to DNN intermediate output.
This can, \forexample, be used to test or formally verify logical
properties that are formulated on the enriched DNN output.


\bibliographystyle{splncs04}
\bibliography{literature} 

\begin{thebibliography}{10}
\providecommand{\url}[1]{\texttt{#1}}
\providecommand{\urlprefix}{URL }
\providecommand{\doi}[1]{https://doi.org/#1}

\bibitem{arrieta_explainable_2020}
Arrieta, A.B., Rodr{\'i}guez, N.D., Ser, J.D., Bennetot, A., Tabik, S.,
  Barbado, A., Garc{\'i}a, S., {Gil-Lopez}, S., Molina, D., Benjamins, R.,
  Chatila, R., Herrera, F.: Explainable artificial intelligence ({{XAI}}):
  {{Concepts}}, taxonomies, opportunities and challenges toward responsible
  {{AI}}. Information Fusion  \textbf{58},  82--115 (2020)

\bibitem{bau_network_2017}
Bau, D., Zhou, B., Khosla, A., Oliva, A., Torralba, A.: Network dissection:
  {{Quantifying}} interpretability of deep visual representations. In: Proc.
  2017 {{IEEE Conf}}. {{Comput}}. {{Vision}} and {{Pattern Recognition}}. pp.
  3319--3327. {IEEE Computer Society} (2017). \doi{10.1109/CVPR.2017.354}

\bibitem{debrabandere_human_2017}
De~Brabandere, S.: Human body ratios. Scientific American (Bring Science Home)
  (Mar 2017),
  \url{https://www.scientificamerican.com/article/human-body-ratios/}

\bibitem{esser_disentangling_2020}
Esser, P., Rombach, R., Ommer, B.: A disentangling invertible interpretation
  network for explaining latent representations. In: Proc. 2020 {{IEEE Conf}}.
  {{Comput}}. {{Vision}} and {{Pattern Recognition}}. pp. 9220--9229 (Jun 2020)

\bibitem{fong_net2vec_2018}
Fong, R., Vedaldi, A.: {{Net2Vec}}: {{Quantifying}} and explaining how concepts
  are encoded by filters in deep neural networks. In: Proc. 2018 {{IEEE Conf}}.
  {{Comput}}. {{Vision}} and {{Pattern Recognition}}. pp. 8730--8738. {IEEE
  Computer Society} (2018)

\bibitem{geirhos_imagenettrained_2019}
Geirhos, R., Rubisch, P., Michaelis, C., Bethge, M., Wichmann, F.A., Brendel,
  W.: {{ImageNet}}-trained {{CNNs}} are biased towards texture; increasing
  shape bias improves accuracy and robustness. In: Proc. 7th {{Int}}. {{Conf}}.
  {{Learning Representations}}. {OpenReview.net} (2019)

\bibitem{ghorbani_automatic_2019}
Ghorbani, A., Wexler, J., Zou, J.Y., Kim, B.: Towards automatic concept-based
  explanations. In: Advances in {{Neural Information Processing Systems}} 32.
  pp. 9273--9282 (2019)

\bibitem{gu_semantics_2019}
Gu, J., Tresp, V.: Semantics for global and local interpretation of deep neural
  networks. CoRR  \textbf{abs/1910.09085} (Oct 2019),
  \url{http://arxiv.org/abs/1910.09085}

\bibitem{he_mask_2017}
He, K., Gkioxari, G., Doll{\'a}r, P., Girshick, R.: Mask {{R}}-{{CNN}}. In:
  2017 {{IEEE International Conference}} on {{Computer Vision}} ({{ICCV}}). pp.
  2980--2988 (Oct 2017)

\bibitem{iso/tc22/sc32_iso_2018a}
{ISO/TC 22/SC 32}: {{ISO}} 26262-6:2018(En): {{Road}} Vehicles \textemdash{}
  {{Functional}} Safety \textemdash{} {{Part}} 6: {{Product}} Development at
  the Software Level, {{ISO}} 26262:2018(En), vol.~6. {International
  Organization for Standardization}, second edn. (Dec 2018)

\bibitem{kim_interpretability_2018}
Kim, B., Wattenberg, M., Gilmer, J., Cai, C., Wexler, J., Viegas, F., Sayres,
  R.: Interpretability beyond feature attribution: {{Quantitative}} testing
  with concept activation vectors ({{TCAV}}). In: Proc. 35th {{Int}}. {{Conf}}.
  {{Machine Learning}}. Proceedings of {{Machine Learning Research}}, vol.~80,
  pp. 2668--2677. {PMLR} (Jul 2018)

\bibitem{kingma_adam_2015}
Kingma, D.P., Ba, J.: Adam: {{A}} method for stochastic optimization. In: Proc.
  3rd {{Int}}. {{Conf}}. {{Learning Representations}} (2015)

\bibitem{krizhevsky_one_2014}
Krizhevsky, A.: One weird trick for parallelizing convolutional neural
  networks. CoRR  \textbf{abs/1404.5997} (2014),
  \url{http://arxiv.org/abs/1404.5997}

\bibitem{larson_standard_2014}
Larson, D.: Standard proportions of the human body (Jan 2014),
  \url{https://www.makingcomics.com/2014/01/19/standard-proportions-human-body/}

\bibitem{lin_microsoft_2014}
Lin, T.Y., Maire, M., Belongie, S.J., Hays, J., Perona, P., Ramanan, D.,
  Doll{\'a}r, P., Zitnick, C.L.: Microsoft {{COCO}}: {{Common}} objects in
  context. In: Proc. 13th {{European Conf}}. {{Computer Vision}} - {{Part V}}.
  Lecture {{Notes}} in {{Computer Science}}, vol.~8693, pp. 740--755. {Springer
  International Publishing} (2014)

\bibitem{olah_feature_2017}
Olah, C., Mordvintsev, A., Schubert, L.: Feature visualization. Distill
  \textbf{2}(11) (Nov 2017). \doi{10.23915/distill.00007}

\bibitem{rabold_expressive_2020}
Rabold, J., Schwalbe, G., Schmid, U.: Expressive explanations of {{DNNs}} by
  combining concept analysis with {{ILP}}. In: {{KI}} 2020: {{Advances}} in
  {{Artificial Intelligence}}. pp. 148--162. Lecture {{Notes}} in {{Computer
  Science}}, {Springer International Publishing} (2020).
  \doi{10.1007/978-3-030-58285-2\_11}

\bibitem{salehi_tversky_2017}
Salehi, S.S.M., Erdogmus, D., Gholipour, A.: Tversky loss function for image
  segmentation using {{3D}} fully convolutional deep networks. In: Proc. 8th
  {{Int}}. {{Workshop Machine Learning}} in {{Medical Imaging}}. vol. 10541,
  pp. 379--387. {Springer} (2017)

\bibitem{schwalbe_concept_2020}
Schwalbe, G., Schels, M.: Concept enforcement and modularization as methods for
  the {{ISO}} 26262 safety argumentation of neural networks. In: Proc. 10th
  {{European Congress Embedded Real Time Software}} and {{Systems}} (Jan 2020),
  \url{https://hal.archives-ouvertes.fr/hal-02442796}

\bibitem{simonyan_very_2015}
Simonyan, K., Zisserman, A.: Very deep convolutional networks for large-scale
  image recognition. In: Proc. 3rd {{Int}}. {{Conf}}. {{Learning
  Representations}} (2015)

\bibitem{yeh_completenessaware_2020}
Yeh, C.K., Kim, B., Arik, S., Li, C.L., Pfister, T., Ravikumar, P.: On
  completeness-aware concept-based explanations in deep neural networks. In:
  Advances in Neural Information Processing Systems 33. vol.~33, pp.
  20554--20565 (2020)

\end{thebibliography}

\end{document}